%% file: acl_latex.tex
\pdfoutput=1

\documentclass[11pt]{article}

\usepackage[final]{acl}

\usepackage{times}
\usepackage{latexsym}
\usepackage{graphicx}
\usepackage{textcomp}
\usepackage{xcolor}
\usepackage{physics}
\usepackage{mathdots}
\usepackage{subfigure}
\usepackage{pgfplots}
\usepackage{tikz}
\usetikzlibrary{patterns}
\usepackage{multirow}
\usepackage{multicol}
\usepackage{array}
\pgfplotsset{compat=1.18}

\usepackage[ruled,linesnumbered]{algorithm2e}
\usepackage{booktabs}

\usepackage{amsmath,amssymb,amsfonts}
\DeclareMathOperator*{\argmax}{arg\,max}

\usepackage[T1]{fontenc}

\usepackage[utf8]{inputenc}

\usepackage{microtype}

\usepackage{inconsolata}

\usepackage{graphicx}

\title{DIDS: Domain Impact-aware Data Sampling for \\ Large Language Model Training}

\author{
 \textbf{Weijie Shi\textsuperscript{1}\thanks{Equal contribution}\thanks{\small{
   \textbf{Email:} \href{mailto:wshiah@connect.ust.hk}{wshiah@connect.ust.hk}
 }}},
 \textbf{Jipeng Zhang\textsuperscript{1}\footnotemark[1]},
 \textbf{Yaguang Wu\textsuperscript{2}},
 \textbf{Jingzhi Fang\textsuperscript{1}},
\\
 \textbf{Shibo Zhang\textsuperscript{2}},
 \textbf{Yao Zhao\textsuperscript{3}},
 \textbf{Hao Chen\textsuperscript{1}},
 \textbf{Ruiyuan Zhang\textsuperscript{1}},
\\
 \textbf{Yue Cui\textsuperscript{1}},
 \textbf{Jia Zhu\textsuperscript{4}},
 \textbf{Sirui Han\textsuperscript{1}\thanks{Corresponding authors}},
 \textbf{Jiajie Xu\textsuperscript{5}}, \textbf{Xiaofang Zhou\textsuperscript{1}\footnotemark[3]}
\\
\\
 \textsuperscript{1}The Hong Kong University of Science and Technology, \textsuperscript{2}MetaX, \\
 \textsuperscript{3}Alibaba Group, 
 \textsuperscript{4}Zhejiang Normal University, \textsuperscript{5}Soochow University
}

\begin{document}
\maketitle
\begin{abstract}
Large language models (LLMs) are commonly trained on multi-domain datasets, where domain sampling strategies significantly impact model performance due to varying domain importance across downstream tasks. Existing approaches for optimizing domain-level sampling strategies struggle with maintaining intra-domain consistency and accurately measuring domain impact. In this paper, we present Domain Impact-aware Data Sampling (DIDS). To ensure intra-domain consistency, a gradient clustering algorithm is proposed to group training data based on their learning effects, where a proxy language model and dimensionality reduction are employed to reduce computational overhead. To accurately measure domain impact, we develop a Fisher Information Matrix (FIM) guided metric that quantifies how domain-specific parameter updates affect the model's output distributions on downstream tasks, with theoretical guarantees. Furthermore, to determine optimal sampling ratios, DIDS combines both the FIM-guided domain impact assessment and loss learning trajectories that indicate domain-specific potential, while accounting for diminishing marginal returns. Extensive experiments demonstrate that DIDS achieves 3.4\% higher average performance while maintaining comparable training efficiency. The code is available at \url{https://github.com/shiweijiezero/DIDS}.
\end{abstract}

\section{Introduction}
Large language models (LLMs) have demonstrated remarkable capabilities across diverse tasks through training on massive multi-domain datasets, enabling robust generalization and adaptation abilities \cite{touvron2023llama,weber2024redpajama,biderman2023pythia,zhang2022opt}. While the composition of training data (e.g., code, scientific papers, web text) significantly shapes model performance, their relative importance varies substantially with respect to target applications. On the one hand, some data domains contribute positively to model performance, whereas others may even impair effectiveness and waste computational resources \cite{xia2024less,zhou2024lima}. On the other hand, each domain's contribution to model learning evolves dynamically throughout the training process \cite{luo2024velocitune,kang2024autoscale}. This necessitates an approach for optimizing domain-level data sampling strategies during LLM training to maximize performance across downstream tasks while maintaining training efficiency. Unfortunately, designing such an algorithm presents several crucial challenges.

\textbf{Intra-domain Consistency.} A fundamental prerequisite for effective domain-level sampling strategies is maintaining data consistency within each domain. Existing approaches either rely on data source categorization \cite{xie2024doremi,fandoge} or employ BERT semantic clustering \cite{fan2024dynamic}. However, these methods often fail to ensure that data within each domain has similar training effects, which is crucial for making domain-level sampling strategies meaningful. To address this limitation, gradient information serves as a more direct measure of training impact. Gradients inherently capture how each data point influences model parameters during training, enabling us to group samples based on their learning effects rather than superficial characteristics.

\textbf{Domain Impact and Mixing Strategy.} The next key challenge lies in accurately measuring each domain's impact on downstream tasks throughout the dynamic training process. Unfortunately, existing grid search methods \cite{ye2024data,liu2024regmix,mckinzie2025mm1} are computationally intensive and cannot adapt to the dynamic of domain importance during training, while gradient similarity approaches \cite{fandoge,fan2024dynamic} only measure the instantaneous parameter update direction alignment without considering how these updates actually affect the model's predictive behavior on downstream tasks. To quantify such influence in a principled way, a natural objective is minimizing the output distributional discrepancy between how different domains' updates shift the model's predictions. Beyond measuring impact, determining optimal sampling ratios requires balancing computation resources across all downstream tasks while considering the marginal utility of domain data, as domain-specific capabilities may saturate with diminishing returns over time.

In this paper, we propose \textbf{D}omain \textbf{I}mpact-aware \textbf{D}ata \textbf{S}ampling (DIDS), which dynamically optimizes domain-level sampling probability by measuring domains' impact on model's predictive behavior. To ensure intra-domain consistency, a gradient clustering algorithm is proposed to group training data, where a small proxy language model is employed instead of the full-size model to reduce computation cost, followed by gradient norm-based subsampling and Johnson-Lindenstrauss random projection for dimensionality reduction. To accurately measure domain impact, a Fisher Information Matrix (FIM) guided metric is developed to quantify the output distributional shift based on the second-order Taylor approximation of KL divergence, enabling efficient assessment of how each domain affects the model's predictive behavior on downstream tasks. We also provide theoretical foundations for the FIM-guided metric. To determine domain sampling proportions, weights are computed by combining both the FIM-guided domain impact on downstream tasks and their loss improvement trajectories indicating learning potential. Extensive experiments on Llama-3.1 across 9 downstream tasks demonstrate that DIDS achieves 3.4\% higher average performance. Our contributions are summarized as follows:
\begin{itemize}
    \item We present a gradient-based data clustering that leverages proxy models and dimensionality reduction to group training samples, ensuring intra-domain training consistency.
    \item We propose a FIM-guided impact metric that measures how domain-specific parameter updates shift model's output distributions on downstream tasks, enabling accurate assessment of domain importance during training with theoretical foundations.
    \item We design DIDS, a domain sampling framework that dynamically adjusts mixing ratios by combining domain impact with learning trajectories, accounting for diminishing marginal returns of domain-specific performance.
\end{itemize}

\section{Related Work}
\subsection{Instance-level Data Sampling}
Instance-level data sampling approaches for language model training primarily focus on selecting high-quality training samples that maximize model performance. LIMA \cite{zhou2024lima} demonstrates that a small set of 1,000 carefully curated prompts can achieve strong performance, \citet{ge2024clustering} ensures both quality and diversity through BERT-based scoring and clustering, and DEITA \cite{liu2023makes} further considers instruction complexity by ChatGPT. Moreover, to align requirements of the specific downstream task, DSIR \cite{xie2023data} utilizes N-gram feature-based importance resampling, while LESS \cite{xia2024less} and TracIn \cite{pruthi2020estimating} leverage gradient-based methods to identify influential training samples through gradient alignment and descent tracing. However, these approaches either lack downstream task awareness or are computationally expensive, motivating domain-level sampling strategies.

\input{figure/architecture}

\subsection{Domain-level Data Sampling}
Domain-level data sampling strategies can be categorized into static and online methods. Static methods determine fixed sampling ratios using proxy models before full-scale training begins. MM1 \cite{mckinzie2025mm1} employs grid search to evaluate different sampling ratios empirically, while Mixing Laws \cite{ye2024data} extends this by proposing scaling law formulas to model the relationship between mixing ratios and model performance. REGMIX \cite{liu2024regmix} introduces regression models to predict this scaling curve. Moreover, Doremi \cite{xie2024doremi} incorporates reference models to consider excess loss, and Doge \cite{fandoge} utilizes gradient alignment between training and validation sets. However, AUTOSCALE \cite{kang2024autoscale} reveals that optimal mixing ratios derived from proxy models may not transfer effectively to larger models. Thus, online methods directly adjust sampling ratios throughout the training process. DGA \cite{fan2024dynamic} extends Doge's gradient-based approach to online scenarios, while Velocitune \cite{luo2024velocitune} monitors learning velocity to adaptively adjust domain proportions. 

Moreover, DRPruning \cite{deng2024drpruning} employs distributionally robust optimization to iteratively shift data distribution toward underperforming domains during training, ensuring balanced recovery across all areas rather than allowing some domains to lag behind after model pruning. It shares our motivation for adaptive domain reweighting but focuses specifically on post-pruning recovery scenarios. DDK \cite{liu2024ddk} computes perplexity ratios between teacher and student models across domains and uses factor-smooth updating mechanisms to periodically adjust sampling probabilities. DDK allocates more training data to domains where the student model underperforms relative to the teacher, thereby reducing performance gaps during knowledge distillation.

Yet existing methods either rely on gradient similarity alone without capturing downstream impact, or use computationally expensive techniques like Scaling Law, limiting their practicality. This motivates our efficient, theoretically grounded approach to dynamic domain-level sampling.

\section{Problem Formulation}
In this part, we formalize the problem of optimizing domain-level sampling strategy for LLM training. 

Consider $\mathcal{D}=\{D_1,...,D_k\}$ denote a training dataset comprising $k$ disjoint domains and $\mathcal{S} = \{S_1, ..., S_m\}$ represent a collection of downstream tasks. Given a large language model $f_\theta$ parameterized by $\theta$ and a computational budget of $n$ training instances, our goal is to optimize the model's performance across all tasks by adjusting the sampling probabilities across different domains during parameter training.

We characterize the domain sampling strategy through a probability vector $\mathbf{p}_t = [p_{t,1}, ..., p_{t,k}]$ at each training step $t$, where $p_{t,i}$ represents the sampling probability from domain $D_i$ subject to the $(k-1)$-dimensional probability simplex $\Pi^{k-1}$:
\begin{equation} 
    \mathbf{p}_t \in \Pi^{k-1} = \{p_{t,i} \geq 0, \sum_{i=1}^k p_{t,i} = 1\}
\end{equation}

The objective of the training process follows a bi-level optimization framework to optimize both model parameters $\theta$ and sampling probabilities $\mathbf{p}$:
\begin{equation}
    \max_{\theta, \mathbf{p} \in \Pi^{k-1}} \sum_{j=1}^m \text{Acc}_j(f_\theta; S_j)
\end{equation}
where $\text{Acc}_j(f_\theta; S_j)$ measures the model’s accuracy on downstream task $S_j$.

To update the model parameters, we perform standard gradient descent:
\begin{equation}
    \theta_{t+1} = \theta_t - \eta \nabla \ell(\theta_t, \mathcal{B}_t), \quad \mathcal{B}_t \sim \mathbf{p_t}
\end{equation}
where $\mathcal{B}_t$ denotes a mini-batch sampled according to the domain sampling probabilities $\mathbf{p}_t$, $\eta$ denotes the learning rate, and $\nabla \ell$ computes the loss gradients with respect to the model parameters.

To update the domain sampling probabilities, we periodically adjust the sampling distribution every $\tau$ steps to optimize the expected model performance across all downstream tasks:
\begin{equation}
    \mathbf{p}_{t} = \argmax_{\mathbf{p} \in \Pi^{k-1}} \sum_{j=1}^m \text{Acc}_j(f_{\theta_{t+\tau}}; S_j)
\end{equation}
where $\theta_{t+\tau}$ represents the model parameters after $\tau$ steps of training using sampling distribution $\mathbf{p}_{t}$.


\section{Methodology}

\subsection{Gradient-based Domain Repartition}
Effective domain-level sampling strategies require consistent training behavior within each domain. Traditional approaches to domain partitioning typically rely on superficial characteristics, such as data sources or semantic similarity measured through BERT embeddings. However, these methods often fail to capture how different training samples actually influence model learning. For instance, mathematical proofs and programming implementations, despite being traditionally categorized into different domains, often induce similar gradient patterns during training due to their shared logical reasoning nature. Conversely, two web documents from the same domain might trigger drastically different parameter updates. To better organize the training data, a gradient-based domain repartitioning is suitable to directly reflect parameter update behaviors. 

Unfortunately, computing and clustering gradients using a full-size LLM for all samples would be computationally prohibitive. A small proxy model maintaining the same architecture but with reduced width and depth serves as an efficient alternative. For each training sample $x_i \in \mathcal{D}$, gradient computation yields vector $g_i$ through
$\nabla \ell(\theta', x_i)$. Here we only keep the last 10\% gradients to accelerate computation. To make clustering computationally feasible, gradient norm-based subsampling retains only the top-k elements with the largest magnitudes in each gradient vector. Next, dimensionality reduction is performed via Johnson-Lindenstrauss random projection \cite{park2023trak} to compress the gradient vectors from parameter-scale dimensionality (millions-level) to a clustering-manageable dimension (thousands-level):
\begin{equation}
\tilde{g}_i = R^T g_i, \quad R \in \mathbb{R}^{h \times s}
\end{equation}
where $h$ represents the original dimension and $s$ denotes the target dimension satisfying $s \ll h$. The random projection matrix $R$ is initialized by randomly orthogonal matrices. The detailed Johnson-Lindenstrauss theorem and initialization methods are provided in Appendix A.

Building upon initial semantic categorization, k-means clustering on these reduced gradient vectors refines each domain, where the number of clusters serves as a hyperparameter. The resulting domains are denoted as $\mathcal{D} = \{D_1, ..., D_k\}$, where $k$ represents the total number of domains.

\subsection{FIM-guided Domain Impact}
After establishing consistent domain partitions, a key challenge is accurately measuring how each domain's training data impacts model performance on downstream tasks. Existing approaches either rely on computationally expensive grid search methods that cannot adapt to dynamic training processes, or use gradient similarity metrics. For example, DGA \cite{fan2024dynamic} measures the domain impact on specific downstream tasks as:
\begin{equation}
    I(D_i,S_j) = \mathbb{E}_{x_i\sim D_i, x_j\sim S_j} [\langle\nabla\ell(\theta^t, x_i), \nabla\ell(\theta^t, x_j)\rangle]
\end{equation}
where $I(D_i,S_j)$ measures the impact of training domain $D_i$ on downstream task $S_j$, $\langle\nabla\ell(\theta^t, x_i), \nabla\ell(\theta^t, x_j)\rangle$ represents the inner product of gradients. However, they only capture instantaneous parameter update directions without considering their actual effects on model behavior. We need a more principled approach that can efficiently quantify how domain-specific parameter updates influence the model's predictive distributions on target tasks. 

To this end, we propose a Fisher Information Matrix (FIM) guided metric that quantifies the output distributional changes induced by domain-specific data. The core insight is that the Kullback-Leibler (KL) divergence between the original and updated model predictions provides a natural measure of how parameter updates affect model behavior. Due to the intractable nature of direct KL divergence computation in infinite input spaces, here we employ a second-order Taylor approximation. 

For notational simplicity, let $p(y|\theta)$ be denoted as $p(\theta)$, $\theta_{D_i} = \theta + \nabla \ell_{D_i}$ and $\theta_{S_j} = \theta + \nabla \ell_{S_j}$ represent the parameters after updates from domain $D_i$ and task $S_j$ respectively, and $\Delta = \nabla \ell_{S_j}-\nabla \ell_{D_i}$ represent the gradient difference between downstream task updates and training domain updates. Formally, we define the domain impact metric as:
\begin{equation}
\begin{split}
    I(D_i,S_j) &= \text{KL}[p(\theta_{D_i}) \parallel p(\theta_{S_j})] \\
    &= \int p(\theta_{D_i})\log \frac{p(\theta_{D_i})}{p(\theta_{S_j})}dy \\
    &= \mathbb{E}_{p(\theta_{D_i})}[\log p(\theta_{D_i})]\!\! -\! \! \mathbb{E}_{p(\theta_{D_i})}\![\log p(\theta_{S_j})]
\end{split}
\end{equation}
When the gradient updates are small (i.e., $\nabla \ell_{D_i} \approx \nabla \ell_{S_j} \approx 0$), we can approximate using second-order Taylor expansion around $\theta_{D_i}$ as:
\begin{equation}
\begin{split}
    &\text{KL}[p(\theta_{D_i}) \!\! \parallel \! p(\theta_{S_j})] \! \approx \mathbb{E}_{p(\theta_{D_i})}[\log p(\theta_{D_i})] \!-\! \mathbb{E}_{p(\theta_{D_i})}[ \\
    &\quad \log p(\theta_{D_i}) \!+\! \nabla\!\log p(\theta_{D_i})\Delta \!+\! \frac{1}{2}\!\Delta\!^T\nabla^2\!\log p(\theta_{D_i})\Delta] \\
    &= -\mathbb{E}_{p(\theta_{D_i})}[\nabla\log p(\theta_{D_i})\Delta] \\
    &\quad - \mathbb{E}_{p(\theta_{D_i})}\left[\frac{1}{2}\Delta^T\nabla^2\log p(\theta_{D_i})\Delta\right]
\end{split}
\end{equation}
The first term can be simplified through integration-differentiation interchange:
\begin{equation}
\begin{split}
    \mathbb{E}_{p(\theta_{D_i})}[\nabla\!\log p(\theta_{D_i})\Delta]\! &=\! \! \int_{\theta_{D_i}} \!\!\!\! \frac{\nabla p(\theta_{D_i})}{p(\theta_{D_i})} \! p(\theta_{D_i})\Delta d\theta\!_{D_i} \\
    &=\! \nabla\int_{\theta_{D_i}} p(\theta_{D_i})d\theta_{D_i} \cdot \Delta \\
    &=\! \nabla(1) \cdot \Delta = 0
\end{split}
\end{equation}
For the second term, the expected Hessian of the negative log-likelihood is equivalent to Fisher Information Matrix:
\begin{equation}
\begin{split}
    \mathbb{E}_{p(\theta_{D_i})} [\nabla^2\log p(\theta_{D_i})] &= \mathbb{E}_{p(x|\theta_{D_i})}[\text{H}_{\log p(x|\theta_{D_i})}] \\
    &= -\text{F}
\end{split}
\end{equation}
Considering that the FIM for LLMs is extremely large and cannot be computed at $\theta_{D_i}$ since the model has not been updated, we instead use diagonal approximation at $\theta$ in practice:
\begin{equation}
    \text{F} \approx \mathbb{E}[\nabla\log p(\theta) \odot \nabla\log p(\theta)]
\end{equation}
Note that FIM only measures the local geometry of the parameter space, and the difference between using FIM at $\theta_{D_i}$ and $\theta$ is negligible when the gradient updates are small. Afterward, the domain impact metric could be rewritten as:
\begin{equation}
\begin{split}
    I(D_i,S_j) &= \! \text{KL}[p(\theta_{D_i}) \parallel p(\theta_{S_j})] \\
    &= \! -\mathbb{E}_{p(\theta_{D_i})}\!\left[\frac{1}{2}\Delta^T\nabla^2\log p(\theta_{D_i})\Delta\!\right] \\
    &= \! \frac{1}{2}\Delta^T F \Delta
\end{split}
\end{equation}
This quadratic form captures how the difference in gradient updates affects the model's output distribution, weighted by the FIM which characterizes the local geometry of the parameter space. The complexity analysis is provided in Section \ref{cost}.

\subsection{Dynamic Domain Sampling Strategy}
Building upon the FIM-guided domain impact measurement, a dynamic sampling strategy is proposed to optimize domain mixing ratios by considering both current learning progress and future potential. The sampling probability for each domain is updated periodically using a combination of three key components:

\textbf{Current Performance Impact.} To identify valuable domains that can achieve larger performance improvements with lower sampling probabilities, we compute a utility score for each domain $D_i$ and downstream task $S_j$ that measures the domain's effectiveness in improving task performance:
\begin{equation}
U(D_i, S_j) = \frac{I(D_i, S_j) \cdot l_c}{p_{t-1,i}}
\end{equation}
where $I(D_i, S_j)$ is the normalized FIM-guided impact score, $l_c$ represents the loss improvement on task $S_j$ between consecutive updates $\Delta L(S_j)$, and $p_{t-1,i}$ is the previous sampling probability for domain $D_i$.

\textbf{Future Potential Estimation.} To account for the diminishing returns in domain-specific learning and prioritize unsaturated domains, we introduce a potential factor $l_p$ that estimates future improvement opportunities. Given the loss history ${l_1, ..., l_t}$ for each downstream task, we fit an exponential decay model\footnote{\url{https://scikit-learn.org/stable/auto_examples/linear_model/plot_bayesian_ridge_curvefit.html}}, which is a typical pattern for learning curves:
\begin{equation}
l_t = ae^{-bt} + c
\end{equation}
where parameters $a$, $b$, and $c$ are estimated using curve fitting. The potential factor $l_p$ is then computed as the difference between current loss and predicted future loss:
\begin{equation}
l_p = l_t - l_{t + \tau}
\end{equation}
where $\tau$ represents the prediction window size.

\textbf{Sampling Probability Update.} The final sampling probabilities are updated using an exponential moving average (EMA) to maintain stability:
\begin{equation}
p_{t,i} = \beta p_{t-1,i} + (1-\beta)\left(\frac{\sum_j I(D_i, S_j) \cdot (l_c + l_p)}{p_{t-1,i}}\right)
\end{equation}
where $\beta$ is the EMA momentum coefficient, $l_c$ represents the current loss improvement, and $l_p$ is the estimated potential factor. A softmax normalization ensures valid probability distribution while the division by previous probabilities implements importance sampling correction. The complete algorithm is summarized in Appendix \ref{appendix D}.

\input{figure/tab-q1}
\section{Experiments}
\subsection{Experimental Setup}
\subsubsection{Datasets and Tasks}
We utilize the Tulu-3 \cite{lambert2024t} post-training dataset containing 939,344 samples from 18 sources across web text, academic papers, code, mathematics, and books. The downstream evaluation suite comprises: BIG-Bench Hard (BBH) \cite{suzgun2022challenging} for reasoning and problem-solving, BoolQ \cite{clark2019boolq} for reading comprehension and binary question answering, GSM8K \cite{cobbe2021training} and Minerva-MathQA \cite{lewkowycz2022solving} for mathematical reasoning, IFEval \cite{zhou2023instruction} for instruction following, MMLU \cite{hendrycks2020measuring} for multitask language understanding, PIQA \cite{bisk2020piqa} for physical commonsense reasoning, PubMedQA \cite{jin2019pubmedqa} for biomedical question answering, and TruthfulQA \cite{lin2021truthfulqa} for measuring truthfulness in model responses.

\subsubsection{Baselines}
We evaluate DIDS against several domain sampling strategies: Uniform and Random sampling, Doremi \cite{xie2024doremi}, Velocitune \cite{luo2024velocitune}, Doge \cite{fandoge}, and DGA \cite{fan2024dynamic}. For all baseline implementations, we partition a small subset from the downstream task's validation set to serve as observable samples for domain reweighting. Detailed implementations are provided in Appendix \ref{appendix_C}.

\subsection{Main Results}
Table \ref{tab:main_results} presents comprehensive evaluation results across nine downstream tasks under both multi-task and single-task optimization scenarios. For reference, we include results from the base Llama-3.1-8B model and its variant trained on the full 929k samples. 

For multi-task optimization, DIDS with only 100k samples achieves an average score of 62.3, significantly outperforming all baseline methods while surpassing the performance of full data training at 61.2. Specifically, DIDS improves over the strongest baseline Doge by 2.1 on average, with particularly notable gains on mathematical reasoning tasks such as Minerva-MathQA improving by 2.7 points from 17.8 to 20.5. This demonstrates DIDS's effectiveness in identifying and prioritizing the most impactful training samples across diverse downstream tasks. Notably, we observe that for some tasks like MMLU and PIQA where the base model is already approaching saturation, additional training with irrelevant data can be detrimental, as evidenced by the Full Data approach's performance decline from 64.7 to 64.3 on MMLU. Furthermore, given the limited training data budget, unbalanced resource allocation across multiple tasks can lead to improved performance on some tasks at the expense of others, as demonstrated by DGA's poor performance of 42.1 on IFEval.

When optimizing for individual tasks, DIDS demonstrates even stronger performance with an average score of 63.7, surpassing the second-best method DGA by 2.1. DIDS shows significant gains on Knowledge-intensive tasks, with IFEval increasing from 53.2 to 57.5 and TruthfulQA improving from 38.5 to 44.8. This indicates that DIDS's FIM-guided domain impact measurement and dynamic sampling strategy are especially effective when focusing on specific downstream objectives. Notably, even with just 100k samples, roughly 10 percent of the full dataset, DIDS achieves higher average performance than training on the full 929k samples with scores of 63.7 versus 61.2.

\input{figure/tab-q2}
\subsection{Ablations}
To analyze the contribution of each component in DIDS, we conduct ablation experiments by progressively removing key components through gradient-based clustering DIDS-GC, FIM-guided impact measurement DIDS-FIM, and loss trajectory consideration DIDS-LT. Results are shown in Table \ref{tab:ablation}. DIDS-GC replaces gradient-based clustering with BERT semantic clustering, leading to a 1.8-point drop in average performance from 46.9 to 45.1. DIDS-FIM removes the FIM-guided impact measurement, causing a 2.9-point decline to 44.0, most notably affecting TruthfulQA with a 4.5-point drop and IFEval with a 3.7-point decrease. DIDS-LT eliminates the loss trajectory and saturation consideration, resulting in 2.8-point decrease to 44.1, demonstrating that dynamic adaptation to learning progress is crucial for optimal performance. These results show that each component contributes significantly to DIDS effectiveness.

\input{figure/tab-q3}
\subsection{Efficiency Analysis} \label{cost}
To comprehensively evaluate DIDS's computational overhead, we analyze the efficiency of each component: gradient-based clustering, FIM-guided impact measurement, and loss trajectory estimation. Our implementation optimizes computational costs by retaining gradients only from the final 10\% of layers, requiring complete forward passes but partial backward passes. Table \ref{tab:efficiency} presents the computational requirements in terms of TFLOPs and GPU Hours on H800.

Base training of an 8B parameter model on 1B tokens requires $5.47 \times 10^4$ TFLOPs for forward and backward passes, consuming approximately 101.6 GPU hours. For the clustering component processing 1B tokens, we evaluate two approaches using 500M models. BERT semantic clustering requires only forward passes at $7.77 \times 10^2$ TFLOPs, while gradient-based clustering with dimensionality reduction necessitates both forward and partial backward computation at $1.87 \times 10^3$ TFLOPs, requiring 1.5 and 3.3 GPU hours respectively.

For domain impact measurement using an 8B parameter base model with 25 mixing ratio updates, we compare FIM-guided metrics against gradient alignment. Across 72 training domains, maintaining running averages of domain-specific gradients incurs negligible overhead. Evaluating 9 downstream tasks with 200 samples per task, gradient alignment requires $9.86 \times 10^1$ TFLOPs. DIDS additionally computes FIM diagonal elements, adding negligible overhead at approximately $1.78 \times 10^2$ TFLOPs, totaling 0.2 GPU hours. The loss trajectory estimation component introduces minimal computational burden below $10^{-1}$ TFLOPs as it only involves scalar loss value curve fitting. While DIDS introduces roughly 1.9\% additional computational cost compared to DGA, this overhead is justified by substantial performance improvements and reduced training data requirements.

\subsection{Parameter Analysis}
\input{figure/fig-q4}

\subsubsection{Impact of Update Frequency}
Figure \ref{fig:parameter-analysis}a shows how the number of domain sampling probability updates during training affects model performance. When using only 5 updates throughout the entire training process, DIDS achieves an average score of 58.2, which is comparable to the random sampling baseline at 58.9. As we increase the number of updates to 25 and 45, DIDS shows substantial improvements, reaching scores of 60.1 and 61.8 respectively. The performance continues to improve with 65 updates, achieving 62.3, and peaks at 62.4 with 85 updates. However, further increasing to 95 updates leads to a slight performance decline back to 62.3. DGA exhibits a similar trend but with lower overall performance, reaching its peak of 60.1 at 65 updates. Random sampling maintains a constant performance of 58.9 regardless of update frequency, serving as a stable baseline. These results suggest that performing a limited update number during training provides optimal performance for domain sampling strategies.

\subsubsection{Impact of Irrelevant Data Ratio}
To evaluate DIDS's robustness to noise in training data, we introduce varying proportions of irrelevant financial domain data and measure the model performance. As shown in Figure \ref{fig:parameter-analysis}b, DIDS demonstrates strong resilience to irrelevant data. Starting at a baseline performance of 62.3 with no irrelevant data, DIDS maintains and even improves its performance as noise increases, reaching a peak of 63.5 at 20\% irrelevant data before showing slight decline to 63.1 at 25\%. In contrast, both comparison methods exhibit clear degradation with increased noise. DGA's performance drops from 58.5 to 57.1, showing moderate sensitivity to irrelevant data. Random sampling demonstrates the most severe degradation, falling from 58.9 to 54.2. These results highlight DIDS's robust ability to identify and leverage valuable training samples through its FIM-guided impact measurement, even in challenging scenarios with substantial noise in the training dataset.

\input{figure/fig-q5}
\subsubsection{Impact of Proxy Model Size}
We evaluate DIDS using different sizes of proxy models for gradient-based clustering: 500M, 1B, and the full 8B target model. As shown in Figure \ref{fig:model-analysis}a, the choice of proxy model size has minimal impact on final performance, with average scores of 62.3, 62.4, and 62.5 respectively. This validates our design choice of using a 500M proxy model for clustering, as it provides comparable quality while significantly reducing computational costs.

\subsubsection{Impact of Domain Partition Count}
We further examine how the granularity of domain partitioning affects model performance. Figure \ref{fig:model-analysis}b shows that increasing domains from the initial 18 based on data sources leads to substantial early improvements in performance. The average score rises sharply from 61.4 to 62.0 when increasing to 36 domains, followed by moderate gains up to 62.3 with 72 domains. However, further partitioning yields small returns, with performance plateauing around 62.7 even when scaling to 1152 domains. Based on this analysis, we select 72 domains as our default configuration to balance effectiveness and computational efficiency.

\section{Conclusion}
In this paper, we proposed DIDS, a domain impact-aware data sampling framework for large language model training. To ensure consistent domain partitioning, DIDS groups training samples based on gradient patterns, which leads to more effective sampling decisions. FIM-guided metrics measure domain impact accurately, while dynamic sampling optimization combines impact assessment with learning trajectories. Experiments demonstrated that DIDS achieves superior performance across multiple tasks using only 10\% training data.

\section*{Limitations}
Our work has several limitations that should be acknowledged:

First, while DIDS demonstrates strong performance with limited training data, the gradient-based domain repartitioning introduces additional computational overhead when processing large-scale datasets. Although we mitigate this through proxy models and dimensionality reduction, the clustering process still requires considerable computational resources when scaling to billions of training samples. Future work could explore more efficient methods for gradient-based domain partitioning to further reduce this overhead while maintaining clustering quality.

Second, the effectiveness of our FIM-guided impact measurement depends on the accuracy of the diagonal approximation of the Fisher Information Matrix. While this approximation is computationally necessary, it may not capture all parameter interactions, potentially leading to suboptimal sampling decisions in cases where off-diagonal elements are significant. Additionally, our current approach to loss trajectory modeling assumes exponential decay patterns which may not hold for all learning scenarios.

\section*{Ethics Statement}
While DIDS improves training efficiency through selective sampling, it may inadvertently amplify existing biases in the training data by preferentially selecting certain domains based on their measured impact. This could lead to underrepresentation of minority groups or less common topics in the final model. In future applications, DIDS should be integrated with ethical auditing tools to ensure fairness in the sampling process and maintain model ethics.

\section*{Acknowledgments}
We would like to specially thank the support from the A3 project of the HKUST \& MetaX Joint Laboratory. The research work described in this paper was supported by Hong Kong Research Grants Council (grant\# 16202722, 16210625, T43-513/23-N, T22-607/24N). It was partially conducted in JC STEM Lab of Data Science Foundations funded by The Hong Kong Jockey Club Charities Trust. We acknowledge the support of Natural Science Foundation of Zhejiang Province under Grant (LY23F020010).

\bibliography{custom}

\appendix

\section{Johnson-Lindenstrauss Theorem and Random Projection Initialization}

\subsection{Johnson-Lindenstrauss Lemma}
The Johnson-Lindenstrauss lemma states that for any set $X$ of $m$ points in $\mathbb{R}^N$ and $0 < \varepsilon < 1$, there exists a linear map $f: \mathbb{R}^N \to \mathbb{R}^n$ where $n > 8\ln(m)/\varepsilon^2$ such that:

$$
(1-\varepsilon)||u-v||^2 \leq ||f(u)-f(v)||^2 \leq (1+\varepsilon)||u-v||^2
$$
where $u,v \in X$. This theorem guarantees that we can project high-dimensional vectors into a lower-dimensional space while approximately preserving their pairwise distances.

\subsection{Gaussian Random Projection}
For practical implementation, we utilize Gaussian random projection matrices which satisfy the following properties:

1. Spherical symmetry: For any orthogonal matrices $A,B \in O(d)$, $RAR^T$ and $RBR^T$ have identical distributions.

2. Orthogonality: The rows of $R$ are approximately orthogonal.

3. Unit length: Each row of $R$ is normalized to unit length.

The projection matrix $R \in \mathbb{R}^{h \times s}$ is constructed as follows:

1. Generate entries $R_{ij}$ independently according to:
   $$R_{ij} = \begin{cases} 
   +1/\sqrt{t} & \text{with probability } 1/2 \\
   -1/\sqrt{t} & \text{with probability } 1/2
   \end{cases}$$
   where $t = \Omega(k/\varepsilon^2)$ for dimension reduction parameter $k$ and error tolerance $\varepsilon$.

2. Normalize each column to ensure unit length: $\tilde{R}_j = R_j/||R_j||_2$

\subsection{Application to Gradient Dimensionality Reduction}
In the context of gradient-based domain repartitioning, given gradient vectors $g_i \in \mathbb{R}^h$, we project them to $\tilde{g}_i \in \mathbb{R}^s$ where $s \ll h$ using:

$$\tilde{g}_i = R^T g_i$$

The choice of target dimension $s$ balances computational efficiency with distance preservation, typically set as:

$$s = O(\log(m)/\varepsilon^2)$$

where $m$ is the size of gradient vectors and $\varepsilon$ is the desired distance preservation tolerance (typically 0.1-0.3).

This projection enables efficient clustering of gradient vectors while maintaining their essential geometric relationships, facilitating meaningful domain repartitioning based on training behavior patterns.

\section{Implementation Details} \label{appendix_C}
\subsection{Training Data Distribution}
\input{figure/t1}
The training dataset consists of 939,344 samples from 18 diverse sources, covering domains including mathematics, coding, instruction following, and general dialogue. The dataset is available at \url{https://huggingface.co/datasets/allenai/tulu-3-sft-mixture}. The largest components are Tulu 3 Persona MATH with 149,960 samples focusing on mathematical reasoning, followed by FLAN v2 with 89,982 samples of general task instructions, and Evol CodeAlpaca with 107,276 coding-related samples. We provide a detailed breakdown of the dataset composition in Table \ref{tab:data_dist}.

\subsection{Model Architecture}
We implement DIDS based on multiple foundation models: Llama-3.1 ($8$B and $70$B variants), Llama-2-$7$B, and Pythia-$6.9$B. For the proxy model, we utilize Qwen-2.5 ($500$M) and Llama-3.2 ($1$B).

\subsection{Baseline Description}
We compare DIDS against the following baseline methods:
\begin{itemize}
    \item \textbf{Uniform sampling}: A basic baseline that assigns equal probabilities to all domains throughout training.
    
    \item \textbf{Random sampling}: Randomly selects domain data at each step without optimization.
    
    \item \textbf{Doremi} \cite{xie2024doremi}: Trains a proxy model using group distributionally robust optimization to produce offline domain weights for resampling.
    
    \item \textbf{Velocitune} \cite{luo2024velocitune}: Dynamically adjusts domain proportions based on learning velocity guided by scaling laws.
    
    \item \textbf{Doge} \cite{fandoge}: Uses bi-level optimization with a proxy model to learn offline domain weights through gradient alignment.
    
    \item \textbf{DGA} \cite{fan2024dynamic}: Employs online gradient alignment to dynamically estimate optimal pre-training data mixtures.
\end{itemize}
For all baselines, we use identical validation set splits from downstream tasks and tune hyperparameters on a separate development set to ensure fair comparison.

To ensure fair comparison across all methods, we adapted the baseline approaches to work with observable downstream tasks. Specifically, \textbf{DoReMi} was originally designed for in-domain scenarios where test sets follow the same distribution as training data. We seamlessly transferred this algorithm to our downstream task settings by computing excess loss over downstream domains. \textbf{VelociTune} was similarly adapted to observe loss over downstream domains for adjusting training data proportions. \textbf{DOGE} and \textbf{DGA} naturally support downstream domain settings as they compute data proportions based on gradient similarity between training and validation sets (observable sets). Importantly, all baseline methods use the same gradient-based domain partitioning strategy as DIDS, ensuring that computational overhead and domain granularity are consistent across comparisons.

\subsection{Training Details}
The training process employs the AdamW optimizer with a learning rate of $5 \times 10^{-4}$ and linear decay scheduling based on Llama-Factory \cite{zheng2024llamafactory} \footnote{\url{https://github.com/hiyouga/LLaMA-Factory}}. We apply gradient clipping at $1.0$ and weight decay at $0.1$. The model processes sequences with a maximum length of $8,192$ tokens and uses a batch size of $128$, distributed across $8$ H800 GPUs. For DIDS-specific configurations, we set the domain update interval $\tau = 4,000$ steps and use an EMA coefficient $\beta = 0.1$. The framework utilizes $72$ domains for gradient-based clustering. Our dimensionality reduction approach first retains the top $10\%$ of elements by magnitude before projecting the gradients to $1,024$ dimensions.

\subsection{Evaluation Details}
We conduct evaluations using the lm-eval-harness platform \cite{eval-harness}\footnote{\url{https://github.com/EleutherAI/lm-evaluation-harness/}}. All tasks are evaluated under a 3-shot setting using the Vllm backend with chat templates applied. Other parameters follow the platform's default configurations.


\section{Algorithm Description} \label{appendix D}
The Domain Impact-aware Data Sampling (DIDS) algorithm is shown in Algorithm \ref{alg:dids}, which begins with initialization and domain repartitioning. Starting with uniform sampling probabilities across k domains, the algorithm employs a proxy model $f'$ to compute gradients for each training sample $x_i$. These gradients undergo TopK selection and Johnson-Lindenstrauss random projection for dimensionality reduction before k-means clustering establishes the k domains $\{D_1,...,D_k\}$.

At intervals of $\tau$ training steps, DIDS performs domain impact assessment and probability updates. For each domain-task pair $(D_i, S_j)$, the algorithm calculates gradient differences $\Delta$ and impact scores $I(D_i,S_j)$ using the FIM-guided metric, while simultaneously fitting exponential decay curves to task loss histories to estimate future potential $L_p(S_j)$ and current improvements $\Delta L(S_j)$. The algorithm then updates sampling probabilities by computing utility scores $U(D_i)$ that combine these impact scores and loss improvements, applying softmax normalization and exponential moving average with coefficient $\beta$.

Between updates, mini-batches are sampled according to current probabilities $\mathbf{p}_t$, with model parameters updated through standard optimization. This design balances theoretical foundations with practical efficiency through its use of proxy models, strategic gradient processing, and periodic updates, enabling effective domain sampling while maintaining computational feasibility.

\section{Extended Experimental Results}
\subsection{Experiments on Additional Models and Datasets}
To validate the effectiveness of DIDS across different model architectures and datasets, we conducted additional experiments using Mixtral-7B \footnote{https://huggingface.co/mistralai/Mistral-7B-v0.1} alongside Llama-3.1-8B, and tested on both Tulu-v3 and the OpenHermes-2.5 \footnote{https://huggingface.co/datasets/teknium/openhermes} datasets. These comprehensive evaluations strengthen our claims regarding DIDS's broad applicability.

\subsubsection{Results on Mixtral-7B with Tulu-v3}
Table \ref{tab:mixtral_tulu} presents the performance of Mixtral-7B when trained on the Tulu-v3 dataset using various sampling strategies. Similar to our findings with Llama-3.1-8B, DIDS demonstrates superior performance across most tasks, achieving an average score of 61.2 in multi-task optimization, which outperforms the full data training (60.4) despite using only 10\% of the training examples. Notably, DIDS shows substantial improvements on mathematical reasoning tasks (MathQA: 17.8 vs. 15.8 for DGA) and truthfulness (TruthfulQA: 50.5 vs. 47.2 for Doge).

\input{figure/mixtral_tulu}

\subsubsection{Results on Llama-3.1-8B with OpenHermes-2.5}
We further evaluated DIDS using the OpenHermes-2.5 dataset, which offers a different distribution of training data compared to Tulu-v3. Table \ref{tab:llama_openhermes} shows that DIDS consistently outperforms baseline methods across various downstream tasks, achieving an average score of 62.7 in multi-task optimization, which is comparable to training on the full dataset (62.4). In single-task optimization, DIDS achieves even better performance with a score of 64.1, demonstrating the effectiveness of our domain-aware sampling approach on different data distributions.

\input{figure/llama_openhermes}

\subsubsection{Results on Mixtral-7B with OpenHermes-2.5}
To further demonstrate the robustness of our approach across different model-dataset combinations, we evaluated Mixtral-7B on the OpenHermes-2.5 dataset. As shown in Table \ref{tab:mixtral_openhermes}, DIDS continues to outperform baseline methods, achieving an average score of 60.1 in multi-task optimization and 61.3 in single-task optimization. These consistent improvements across different models and datasets strongly support the generalizability of our approach.

\input{figure/mixtral_openhermes}

\subsection{Complete Ablation Study on All Downstream Tasks}
Table \ref{tab:full_ablation} presents a comprehensive ablation study of DIDS across all nine downstream tasks. This expanded analysis provides a more detailed understanding of how each component contributes to the overall performance gains.

\input{figure/full_ablation}

The ablation results clearly demonstrate the contribution of each component of DIDS. Gradient-based Clustering significantly improves performance, as replacing it with BERT semantic clustering leads to a 1.2-point drop in average performance from 62.3 to 61.1. This highlights the importance of grouping data based on actual training effects rather than semantic similarity alone. The FIM-guided Impact Measurement proves crucial, with its removal resulting in a 2.1-point decline to 60.2. This component shows particularly notable benefits for TruthfulQA, PubMedQA, and MathQA tasks, confirming that measuring domain impact through output distributional changes provides a more accurate assessment than gradient similarity alone. Finally, Loss Trajectory Consideration plays a vital role, as its elimination causes a 2.0-point decrease to 60.3, with substantial performance drops in instruction following and truthfulness tasks. This demonstrates the importance of accounting for both current learning progress and future potential when determining sampling probabilities.

\subsection{Domain Mixing Analysis}
Understanding how domain weights evolve during training provides valuable insights into DIDS's operation. Table \ref{tab:domain_weights} shows the weight changes for 10 randomly selected domains (out of 256) throughout the training process for both DIDS and DGA.

\input{figure/domain_weights}

This comparison reveals several key differences between DIDS and DGA. DIDS makes more decisive weight adjustments, with stronger amplification of valuable domains like D023 reaching 2.8 compared to DGA's 1.3, while aggressively reducing less useful domains such as D045 to 0.0 versus DGA's 0.1. This decisive resource allocation contributes significantly to DIDS's superior performance. Furthermore, domains like D078 show non-monotonic weight changes in DIDS, demonstrating its ability to adapt to the dynamic importance of domains during training, in contrast to DGA's more gradual and sometimes inconsistent adjustments. DIDS also tends to converge more quickly to stable domain weights, particularly for highly valuable or less useful domains, enabling more efficient training as the optimal sampling distribution is established earlier. Analysis of domain overlap between DIDS and DGA shows approximately 40\% consistency in domain selection, with substantial differences in the remaining 60\%, highlighting the distinct impact assessment approaches of the two methods and explaining their performance differences.

\subsection{Domain Clustering Insights}
Our gradient-based domain clustering revealed several interesting patterns in how training data is organized:
\begin{itemize}
    \item \textbf{Fine-grained Topic Distinction}: With sufficiently large cluster counts (over 1,000), DIDS can distinguish between closely related topics. For example, in scientific data, middle school and high school biology knowledge are clustered separately, reflecting their different training effects on the model.
    
    \item \textbf{Format Sensitivity}: Within the same subject area (e.g., middle school biology), different question formats like multiple-choice and fill-in-the-blank are clustered into separate domains. This suggests that format significantly influences how data affects model learning, beyond just semantic content.
    
    \item \textbf{Cross-domain Similarity}: Some seemingly distinct topics like mathematical proofs and programming implementations are clustered together due to their shared logical reasoning patterns, despite their different semantic categories in traditional domain partitioning.
    
    \item \textbf{Instruction Pattern Recognition}: Data with similar instruction patterns tends to be clustered together regardless of content topic, highlighting the importance of task structure in determining training effects.
\end{itemize}

\section{Theoretical Analysis of FIM-guided Impact Measurement}
The Fisher Information Matrix (FIM) plays a crucial role in DIDS by enabling accurate measurement of how domain-specific parameter updates affect model behavior on downstream tasks. Here, we provide additional theoretical analysis to justify our approach.

\subsection{Relationship to Model Uncertainty}
The FIM is inherently connected to model uncertainty through the Cramér-Rao bound, which establishes that the inverse of FIM provides a lower bound on the covariance of any unbiased estimator of the parameters. In the context of domain impact measurement, this means that parameters with high Fisher Information have a stronger influence on the model's predictive distribution and consequently on task performance.

For a parameter set $\theta$, the Fisher Information Matrix is defined as:

\begin{equation}
F(\theta) = \mathbb{E}_{p(x|\theta)}\left[\nabla_\theta \log p(x|\theta) \nabla_\theta \log p(x|\theta)^T\right]
\end{equation}

When we compute the impact metric between domain $D_i$ and task $S_j$ as $I(D_i, S_j) = \frac{1}{2}\Delta^T F \Delta$, we are effectively measuring the expected change in the model's log-likelihood on task $S_j$ when updated with domain $D_i$ data, weighted by the parameter sensitivity through the FIM.

\subsection{Consistency with KL Divergence}
The KL divergence between two distributions $p(y|\theta_{D_i})$ and $p(y|\theta_{S_j})$ measures how much information is lost when using one distribution to approximate the other. Our use of the second-order Taylor approximation of KL divergence:

\begin{equation}
\text{KL}[p(\theta_{D_i}) \parallel p(\theta_{S_j})] \approx \frac{1}{2}\Delta^T F \Delta
\end{equation}

Captures this information loss efficiently and accurately when the parameter updates are relatively small. Furthermore, this approximation has the advantage of being positive definite and symmetric (when properly scaled), which makes it a suitable measure for domain impact.

\subsection{Extensions to Alternative Divergences}
While our implementation focuses on KL divergence, the framework can be extended to other divergence measures such as Wasserstein distance or Jensen-Shannon divergence. The general form would remain similar:

\begin{equation}
D[p(\theta_{D_i})\!\! \parallel \!\! p(\theta_{S_j})] \!\! \approx \!\! (\nabla \ell_{S_j} \!- \! \nabla \ell_{D_i})^T M (\nabla \ell_{S_j} \! - \!\nabla \ell_{D_i})
\end{equation}

where $M$ is a metric tensor appropriate for the chosen divergence. This flexibility allows DIDS to be adapted to different notions of distribution similarity based on specific requirements.

\section{Practical Guidelines for DIDS Implementation}
Based on our experiments and analyses, we provide the following practical guidelines for implementing DIDS effectively:

\begin{itemize}
    \item \textbf{Domain Count Selection}: Start with a medium number of domains (approximately 50-100) for gradient-based clustering. Increasing the domain count beyond 100 provides diminishing returns in most cases, while increasing computational cost.
    
    \item \textbf{Update Frequency}: Set the domain sampling probability update interval to approximately 5-10\% of the total training steps. More frequent updates can cause instability, while less frequent updates may miss important adaptation opportunities.
    
    \item \textbf{EMA Coefficient Tuning}: Use an EMA coefficient ($\beta$) of 0.1-0.3 for stability. Lower values allow for more rapid adaptation, which is beneficial in early training stages, while higher values provide stability in later stages.
    
    \item \textbf{Proxy Model Selection}: A proxy model with 5-10\% the size of the target model typically provides a good balance between computational efficiency and gradient similarity. The proxy model should maintain the same architecture family as the target model for best results.
    
    \item \textbf{Downstream Task Selection}: Include a diverse set of downstream tasks in the observation set, covering different capability areas like reasoning, knowledge, instruction following, etc. This diversity ensures balanced optimization across different aspects of model performance.
    
    \item \textbf{FIM Computation Efficiency}: Compute the diagonal FIM approximation using a small batch size (16-32) for efficiency without significant loss in accuracy. The FIM computation only needs to be performed during domain sampling probability updates.
\end{itemize}

\section{Mixed Ratio Analysis}
To further validate the effectiveness of different domain mixing strategies, we conducted a grid search analysis similar to that reported in previous work \cite{zhang2022opt,mckinzie2025mm1}. Table \ref{tab:grid_search} presents results for different mixing ratios of code/math domains versus general domains.

\input{figure/grid_search}

This analysis demonstrates several key points. First, DIDS's dynamic approach outperforms all static mixing ratios across all three tasks, highlighting the limitations of fixed domain proportions throughout training. Second, different tasks show different optimal static mixing ratios - GSM8K and HumanEval benefit from higher proportions of code and math content, while MT-Bench performs better with more balanced or general-leaning distributions. Third, increasing the proportion of specialized domains like code and math significantly improves performance on related tasks such as GSM8K and HumanEval but can negatively impact general capabilities measured by MT-Bench. DIDS effectively navigates these trade-offs through dynamic adaptation.

These results align with findings from industry practices in models like MM1 \cite{mckinzie2025mm1} and Llama3 \cite{touvron2023llama}, where mixed ratios are carefully tuned through extensive grid search. DIDS automates this process and improves upon static optimal ratios through dynamic adaptation.

\section{Domain Partitioning Robustness Analysis}
To evaluate DIDS's sensitivity to domain partitioning quality, we conducted destructive experiments by artificially corrupting the domain structure. Starting with our standard 72-domain gradient-based partitioning, we randomly swapped half of each domain's data with samples from other domains, creating mixed domains that violate intra-domain consistency assumptions.

\begin{table*}[h]
\centering
\small
\begin{tabular}{lccccccccc|c}
\toprule
\textbf{Method} & \textbf{BBH} & \textbf{BoolQ} & \textbf{GSM8K} & \textbf{MathQA} & \textbf{IFEval} & \textbf{MMLU} & \textbf{PIQA} & \textbf{PubMedQA} & \textbf{TruthfulQA} & \textbf{Avg} \\
\midrule
Random & 67.4 & 85.6 & 58.9 & 11.4 & 48.2 & 64.0 & 82.0 & 77.4 & 31.5 & 58.9 \\
DIDS (Original) & \textbf{69.2} & \textbf{87.5} & \textbf{63.0} & \textbf{21.5} & \textbf{57.5} & \textbf{65.8} & \textbf{83.0} & \textbf{81.2} & \textbf{44.8} & \textbf{63.7} \\
DIDS (Unreasonable) & 67.4 & 86.1 & 60.6 & 14.5 & 50.6 & 65.1 & 82.6 & 79.6 & 35.4 & 60.2 \\
\bottomrule
\end{tabular}
\caption{Robustness analysis under corrupted domain partitioning on Llama-3.1-8B.}
\label{tab:domain_partition_robustness}
\end{table*}

Table \ref{tab:domain_partition_robustness} demonstrates that while unreasonable partitioning degrades DIDS performance by 3.5 points (from 63.7 to 60.2), it still outperforms random sampling by 1.3 points. This indicates that DIDS exhibits graceful degradation and maintains effectiveness even when domain assumptions are violated, highlighting the robustness of our FIM-guided impact measurement and dynamic sampling components beyond perfect domain organization.

\section{Detailed Computational Cost Analysis}
We provide a comprehensive breakdown of computational costs across all baseline methods to demonstrate DIDS's efficiency advantages.

\textbf{DoReMi and DoGE} employ offline reweighting strategies using 280M parameter proxy models. DoReMi requires: (1) reference model training with uniform sampling ($2.05 \times 10^3$ TFLOPs, 3.8 GPU hours), (2) proxy model training with Group DRO ($2.05 \times 10^3$ TFLOPs, 3.8 GPU hours), and (3) excess loss computation (0.37 TFLOPs, 0.004 GPU hours). DoGE follows similar complexity but uses gradient similarity calculations instead of excess loss computation.

\textbf{Velocitune} employs two phases: (1) target estimation using full 8B models on 51\% subsampled data ($2.79 \times 10^4$ TFLOPs, 51.8 GPU hours), and (2) velocity-guided training with periodic updates every 150 steps ($1.2 \times 10^2$ TFLOPs, 0.2 GPU hours). The method requires significantly more resources due to full-size model training for target estimation.

\textbf{DGA} uses online gradient alignment with minimal overhead for maintaining running averages of domain-specific gradients.

\begin{table*}[h]
\centering
\small
\begin{tabular}{lccc}
\toprule
\textbf{Method} & \textbf{Total TFLOPs} & \textbf{Overhead} & \textbf{Total GPU Hours} \\
\midrule
Base Training & $5.47 \times 10^4$ & - & 101.6 \\
DGA & $5.56 \times 10^4$ & 1.6\% & 103.2 \\
DIDS & $5.67 \times 10^4$ & 3.7\% & 105.2 \\
DoReMi & $5.88 \times 10^4$ & 7.5\% & 109.2 \\
DoGE & $5.88 \times 10^4$ & 7.5\% & 109.2 \\
Velocitune & $8.27 \times 10^4$ & 51.2\% & 153.6 \\
\bottomrule
\end{tabular}
\caption{Comprehensive computational cost comparison across all methods.}
\label{tab:computational_cost_detailed}
\end{table*}

Table \ref{tab:computational_cost_detailed} demonstrates that DIDS achieves superior performance improvements while maintaining competitive computational efficiency. Our method requires only 3.7\% additional overhead, significantly lower than Velocitune's 51.2\% and comparable to efficient methods like DGA, while providing substantial performance gains.

\section{Out-of-Distribution Generalization Analysis}

For potential overfitting to specific downstream tasks, we evaluate DIDS's generalization capability on unseen out-of-distribution (OOD) tasks across two experimental setups.

For the diverse downstream task setup, we used our standard 9-task evaluation suite (BBH, BoolQ, GSM8K, MathQA, IFEval, MMLU, PIQA, PubMedQA, TruthfulQA) during training, then evaluated on four unseen OOD tasks: WMT16 English-German translation (BLEU), TLDR summarization (Win Rate vs. Llama-3.1-8B), ARC-Challenge science questions (ACC), and MBPP code generation (Pass@1).

For the single task setup, we also tested the extreme case where only one downstream task guides training on MathQA and TruthfulQA.

\begin{table*}[h]
\centering
\small
\begin{tabular}{lccccc}
\toprule
\textbf{Method} & \textbf{WMT16 EN-DE} & \textbf{TLDR} & \textbf{ARC-Challenge} & \textbf{MBPP} & \textbf{Average} \\
\midrule
Llama-3.1-8B (no training) & 17.1 & 50.1 & 76.4 & 55.6 & 49.8 \\
Random (100k) & 16.8 & 47.5 & 82.7 & 57.3 & 51.0 \\
DGA (100k) & 17.3 & 49.8 & 83.6 & 59.7 & 52.6 \\
DIDS (100k) & \textbf{17.2} & \textbf{50.0} & \textbf{85.7} & \textbf{61.3} & \textbf{53.5} \\
\bottomrule
\end{tabular}
\caption{OOD generalization performance with diverse downstream task training.}
\label{tab:ood_diverse}
\end{table*}

\begin{table*}[h]
\centering
\small
\begin{tabular}{lcccc|cccc}
\toprule
 & \multicolumn{4}{c|}{\textbf{Training Target: MathQA}} & \multicolumn{4}{c}{\textbf{Training Target: TruthfulQA}} \\
\textbf{Method} & \textbf{WMT16} & \textbf{TLDR} & \textbf{ARC} & \textbf{MBPP} & \textbf{WMT16} & \textbf{TLDR} & \textbf{ARC} & \textbf{MBPP} \\
\midrule
Llama-3.1-8B (no training) & 17.1 & 50.1 & 76.4 & 55.6 & 17.1 & 50.1 & 76.4 & 55.6 \\
Random & 16.8 & 47.5 & 82.7 & 57.3 & 16.8 & 47.5 & 82.7 & 57.3 \\
DGA & 15.9 & 45.3 & 83.7 & 59.2 & 17.8 & 48.9 & 81.4 & 56.8 \\
DIDS & \textbf{15.9} & \textbf{45.8} & \textbf{86.4} & \textbf{60.6} & \textbf{17.6} & \textbf{54.1} & \textbf{82.3} & \textbf{56.6} \\
\bottomrule
\end{tabular}
\caption{OOD generalization under single-task optimization scenarios.}
\label{tab:ood_single}
\end{table*}

The results in Table \ref{tab:ood_diverse} demonstrate that DIDS maintains competitive OOD performance, especially when optimized for diverse downstream objectives, achieving the highest average performance (53.5) across all OOD evaluation tasks. As shown in Table \ref{tab:ood_single}, when using single downstream tasks, DIDS shows meaningful cross-task transfer—mathematical reasoning benefits scientific reasoning (ARC: 86.4) and code generation (MBPP: 60.6), while truthfulness training improves summarization (TLDR: 54.1). While domain-specific optimization can lead to some specialization effects in unrelated areas, DIDS's FIM-guided approach captures meaningful cross-task dependencies and maintains reasonable generalization capabilities.

\input{figure/alg}

\end{document}

%% file: figure/architecture.tex
\begin{figure*}[t]
\centering
\begin{minipage}[t]{1\linewidth}
\centering
\includegraphics[width=1.0\textwidth]{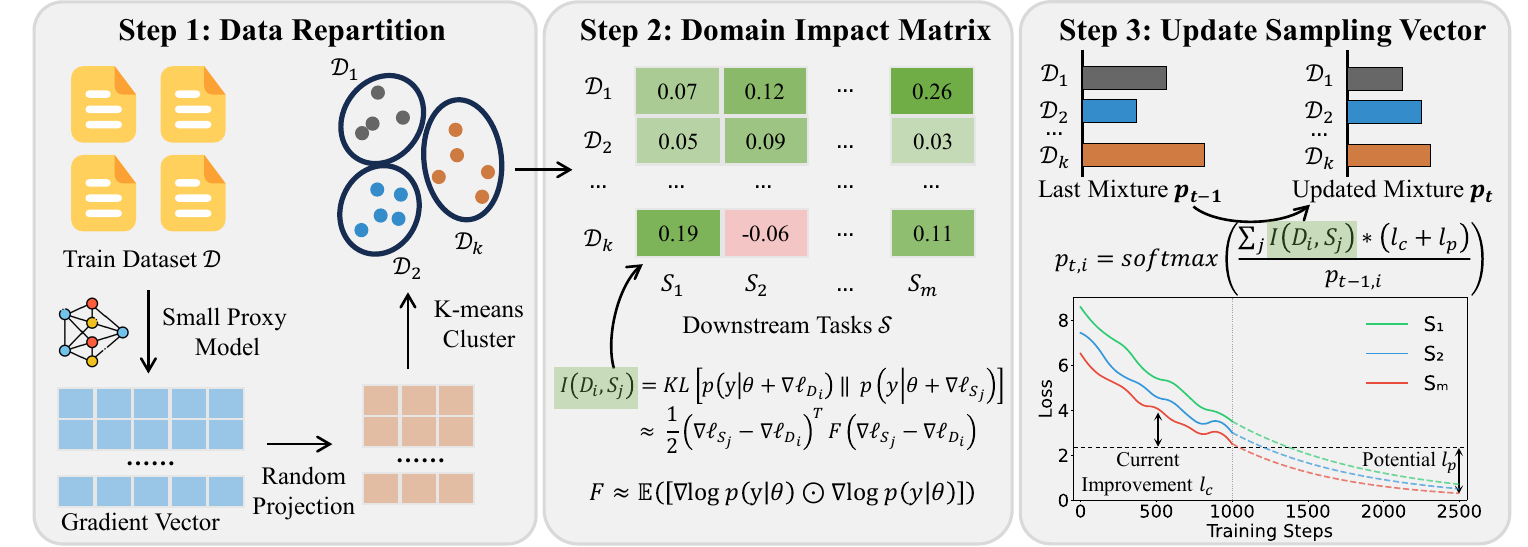}
\end{minipage}
\centering
\caption{Overview of DIDS's three-step process: (1) Domain repartition using gradient-based clustering with a proxy model and dimensionality reduction through random projection, (2) Domain impact measurement using FIM-guided metrics that quantify how domain-specific parameter updates affect model's output distributions on downstream tasks, and (3) Dynamic sampling probability updates that combine both FIM-guided domain impact assessment and loss learning trajectories to account for diminishing marginal returns.}
\label{fig:architecture}
\end{figure*}

%% file: figure/tab-q1.tex
\setlength{\tabcolsep}{4pt}
\begin{table*}[t]
\footnotesize
\begin{center}
\begin{tabular}{l|ccccccccc|c}
\toprule
\multirow{2}{*}{Method} & \multicolumn{2}{c}{Reasoning} & \multicolumn{2}{c}{Mathematics} & \multicolumn{1}{c}{Instruction} & \multicolumn{2}{c}{Commonsense} & \multicolumn{1}{c}{Domain} & \multicolumn{1}{c|}{Truthfulness} & \multirow{2}{*}{Average} \\
\cmidrule(lr){2-3} \cmidrule(lr){4-5} \cmidrule(lr){6-6} \cmidrule(lr){7-8} \cmidrule(lr){9-9} \cmidrule(lr){10-10}
& BBH & BoolQ & GSM8K & MathQA & IFEval & MMLU & PIQA & PubMedQA & TruthfulQA & \\
\midrule
\multicolumn{11}{c}{\textit{Multi-task Optimization}} \\
\midrule
Llama-3.1-8B & \colorbox[rgb]{1.0,0.7,0.7}{62.5} & \colorbox[rgb]{1.0,0.7,0.7}{81.8} & \colorbox[rgb]{1.0,0.7,0.7}{48.9} & \colorbox[rgb]{1.0,0.9,0.9}{15.7} & \colorbox[rgb]{1.0,0.7,0.7}{18.5} & 64.7 & 81.1 & \colorbox[rgb]{1.0,0.8,0.8}{75.8} & \colorbox[rgb]{1.0,0.7,0.7}{28.5} & \colorbox[rgb]{1.0,0.7,0.7}{53.1} \\
+ Full Data (929k) & \colorbox[rgb]{0.8,0.8,1.0}{68.0} & \colorbox[rgb]{0.8,0.8,1.0}{87.3} & \colorbox[rgb]{0.8,0.8,1.0}{65.2} & 16.2 & \colorbox[rgb]{0.8,0.8,1.0}{61.2} & \colorbox[rgb]{1.0,0.85,0.85}{64.3} & \colorbox[rgb]{1.0,0.85,0.85}{81.0} & \colorbox[rgb]{0.85,0.85,1.0}{78.0} & \colorbox[rgb]{1.0,0.8,0.8}{29.5} & \colorbox[rgb]{0.85,0.85,1.0}{61.2} \\
+ Random (100k) & 67.4 & 85.6 & \colorbox[rgb]{1.0,0.8,0.8}{58.9} & \colorbox[rgb]{1.0,0.7,0.7}{11.4} & \colorbox[rgb]{1.0,0.9,0.9}{48.2} & 64.0 & \colorbox[rgb]{0.9,0.9,1.0}{82.0} & 77.4 & 31.5 & 58.9 \\
+ Uniform (100k) & \colorbox[rgb]{1.0,0.8,0.8}{66.2} & \colorbox[rgb]{1.0,0.8,0.8}{83.2} & \colorbox[rgb]{1.0,0.8,0.8}{57.5} & \colorbox[rgb]{1.0,0.8,0.8}{11.8} & 48.2 & 64.1 & 81.5 & \colorbox[rgb]{1.0,0.8,0.8}{76.0} & 31.2 & \colorbox[rgb]{1.0,0.8,0.8}{57.7} \\
\midrule
+ Doremi (100k) & \colorbox[rgb]{0.9,0.9,1.0}{67.5} & 85.8 & 58.8 & 17.5 & 49.8 & 64.5 & 81.9 & 77.8 & 35.8 & 59.9 \\
+ Velocitune (100k) & 67.2 & 85.5 & 56.2 & 17.2 & 49.0 & 64.4 & 81.7 & 77.5 & 35.0 & 59.3 \\
+ Doge (100k) & \colorbox[rgb]{0.85,0.85,1.0}{67.8} & \colorbox[rgb]{0.85,0.85,1.0}{86.0} & \colorbox[rgb]{0.9,0.9,1.0}{57.5} & \colorbox[rgb]{0.85,0.85,1.0}{17.8} & \colorbox[rgb]{0.85,0.85,1.0}{51.2} & 64.6 & \colorbox[rgb]{0.85,0.85,1.0}{82.0} & \colorbox[rgb]{0.9,0.9,1.0}{78.5} & \colorbox[rgb]{0.8,0.8,1.0}{37.2} & \colorbox[rgb]{0.9,0.9,1.0}{60.2} \\
+ DGA (100k) & 67.0 & 85.4 & 58.8 & \colorbox[rgb]{0.8,0.8,1.0}{18.2} & \colorbox[rgb]{1.0,0.7,0.7}{42.1} & \colorbox[rgb]{0.9,0.9,1.0}{64.8} & 81.8 & \colorbox[rgb]{1.0,0.7,0.7}{75.2} & \colorbox[rgb]{1.0,0.8,0.8}{33.4} & \colorbox[rgb]{1.0,0.9,0.9}{58.5} \\
+ DIDS (100k) & \colorbox[rgb]{0.8,0.8,1.0}{68.3} & \colorbox[rgb]{0.85,0.85,1.0}{86.9} & \colorbox[rgb]{0.85,0.85,1.0}{59.0} & \colorbox[rgb]{0.8,0.8,1.0}{20.5} & \colorbox[rgb]{0.9,0.9,1.0}{55.6} & \colorbox[rgb]{0.8,0.8,1.0}{64.9} & \colorbox[rgb]{0.8,0.8,1.0}{82.2} & \colorbox[rgb]{0.8,0.8,1.0}{80.4} & \colorbox[rgb]{0.8,0.8,1.0}{43.0} & \colorbox[rgb]{0.8,0.8,1.0}{62.3} \\
\midrule
\multicolumn{11}{c}{\textit{Single-task Optimization}} \\
\midrule
+ Doremi (100k) & \colorbox[rgb]{0.9,0.9,1.0}{68.8} & 86.2 & 60.8 & 18.2 & 51.2 & 64.8 & \colorbox[rgb]{0.8,0.8,1.0}{82.6} & 78.5 & 37.2 & 60.9 \\
+ Velocitune (100k) & 68.0 & 86.0 & 60.5 & 18.0 & 50.8 & 64.5 & 82.0 & 78.2 & 36.8 & 60.5 \\
+ Doge (100k) & 68.2 & \colorbox[rgb]{0.9,0.9,1.0}{86.8} & 60.9 & 18.4 & 51.5 & 64.9 & 82.2 & \colorbox[rgb]{0.8,0.8,1.0}{79.0} & 37.5 & 61.0 \\
+ DGA (100k) & \colorbox[rgb]{0.85,0.85,1.0}{68.6} & \colorbox[rgb]{0.85,0.85,1.0}{86.5} & \colorbox[rgb]{0.85,0.85,1.0}{61.8} & \colorbox[rgb]{0.85,0.85,1.0}{19.2} & \colorbox[rgb]{0.85,0.85,1.0}{53.2} & \colorbox[rgb]{0.85,0.85,1.0}{65.2} & \colorbox[rgb]{0.85,0.85,1.0}{82.4} & \colorbox[rgb]{0.9,0.9,1.0}{78.8} & \colorbox[rgb]{0.85,0.85,1.0}{38.5} & \colorbox[rgb]{0.85,0.85,1.0}{61.6} \\
+ DIDS (100k) & \colorbox[rgb]{0.8,0.8,1.0}{69.2} & \colorbox[rgb]{0.8,0.8,1.0}{87.5} & \colorbox[rgb]{0.8,0.8,1.0}{63.0} & \colorbox[rgb]{0.8,0.8,1.0}{21.5} & \colorbox[rgb]{0.8,0.8,1.0}{57.5} & \colorbox[rgb]{0.8,0.8,1.0}{65.8} & \colorbox[rgb]{0.8,0.8,1.0}{83.0} & \colorbox[rgb]{0.8,0.8,1.0}{81.2} & \colorbox[rgb]{0.8,0.8,1.0}{44.8} & \colorbox[rgb]{0.8,0.8,1.0}{63.7} \\
\bottomrule
\end{tabular}
\caption{The overall performance comparison. Cells with \colorbox[rgb]{0.9,0.9,1.0}{blue background indicate high scores}, while \colorbox[rgb]{1.0,0.9,0.9}{red background indicates low scores}. The top section shows results when optimizing for multiple downstream tasks simultaneously, while the bottom section shows results when optimizing for individual tasks.}
\label{tab:main_results}
\end{center}
\vspace{-0.6cm}
\end{table*}

%% file: figure/tab-q2.tex
\begin{table}[t]
\footnotesize
\begin{center}
\begin{tabular}{l|cccc|c}
\toprule
Variant & BBH & MathQA & IFEval & TruthfulQA & Avg \\
\midrule
DIDS (100k) & \textbf{68.3} & \textbf{20.5} & \textbf{55.6} & \textbf{43.0} & \textbf{46.9} \\
DIDS-GC & 67.7 & 19.7 & 53.0 & 40.1 & 45.1 \\
DIDS-FIM & 67.2 & 18.6 & 51.9 & 38.5 & 44.0 \\
DIDS-LT & 67.5 & 19.5 & 51.4 & 38.1 & 44.1 \\
\bottomrule
\end{tabular}
\caption{Ablation results. We progressively remove key components: gradient-based clustering (DIDS-GC), FIM-guided impact measurement (DIDS-FIM), and loss trajectory consideration (DIDS-LT).}
\label{tab:ablation}
\end{center}
\vspace{-0.6cm}
\end{table}

%% file: figure/tab-q3.tex
\setlength{\tabcolsep}{4pt}
\begin{table}[t]
\scriptsize
\centering
\begin{tabular}{l|cc|cc}
\toprule
\multirow{2}{*}{Component} & \multicolumn{2}{c|}{TFLOPs} & \multicolumn{2}{c}{GPU Hours} \\
& DGA & DIDS & DGA & DIDS \\
\midrule
Base Training & $5.47 \times 10^{4}$ & $5.47 \times 10^{4}$ & 101.6 & 101.6 \\
Cluster (BERT vs. Gradient) & $7.77 \times 10^{2}$ & $1.87 \times 10^{3}$ & 1.5 & 3.3 \\
Impact (Gradient vs. FIM) & $9.86 \times 10^{1}$ & $1.78 \times 10^{2}$ & 0.1 & 0.2 \\
Loss Trajectory Consideration & - & $< 10^{-1}$ & - & < 0.1 \\
\midrule
Total & $5.56 \times 10^{4}$ & $5.67 \times 10^{4}$ & 103.2 & 105.2 \\
\bottomrule
\end{tabular}
\caption{Computational cost analysis of different components in DIDS. Base training refers to standard training of an 8B parameter model on 1B tokens.}
\label{tab:efficiency}
\vspace{-0.7cm}
\end{table}

%% file: figure/fig-q4.tex
\begin{figure}[t]
    \subfigure[Impact of Update Frequency]{
    \begin{minipage}[t]{0.5\linewidth}
        \centering
        \begin{tikzpicture}[scale=0.45]
        \begin{axis}[
            xlabel={Update Count},
            ylabel={Average Score},
            legend style={at={(0.98,0.02)}, anchor=south east},
            ymajorgrids=true,
            grid style=dashed,
            ymin=55,
            ymax=64,
            font=\Large,
        ]
        \addplot[thick,mark=*,blue] coordinates {
            (5,58.2)
            (25,60.1)
            (45,61.8)
            (65,62.3)
            (85,62.4)
            (95,62.3)
        };
        \addplot[thick,mark=square*,red] coordinates {
            (5,57.8)
            (25,59.2)
            (45,59.8)
            (65,60.1)
            (85,60.0)
            (95,59.9)
        };
        \addplot[thick,mark=triangle*,brown] coordinates {
            (5,58.9)
            (25,58.9)
            (45,58.9)
            (65,58.9)
            (85,58.9)
            (95,58.9)
        };
        \legend{DIDS,DGA,Random}
        \end{axis}
        \end{tikzpicture}
    \end{minipage}
    }%
    \subfigure[Impact of Irrelevant Data]{
    \begin{minipage}[t]{0.5\linewidth}
        \centering
        \begin{tikzpicture}[scale=0.45]
        \begin{axis}[
            xlabel={Irrelevant Data Ratio (\%)},
            ylabel={Average Score},
            legend style={at={(0.98,0.02)}, anchor=south east},
            ymajorgrids=true,
            grid style=dashed,
            ymin=45,
            ymax=66,
            font=\Large,
        ]
        \addplot[thick,mark=*,blue] coordinates {
            (0,62.3)
            (5,62.5)
            (10,62.4)
            (15,63.2)
            (20,63.5)
            (25,63.1)
        };
        \addplot[thick,mark=square*,red] coordinates {
            (0,58.5)
            (5,58.3)
            (10,57.5)
            (15,58.2)
            (20,57.8)
            (25,57.1)
        };
        \addplot[thick,mark=triangle*,brown] coordinates {
            (0,58.9)
            (5,57.8)
            (10,57.4)
            (15,57.0)
            (20,56.4)
            (25,54.2)
        };
        \legend{DIDS,DGA,Random}
        \end{axis}
        \end{tikzpicture}
    \end{minipage}
    }
    \vspace{-0.3cm}
    \caption{Effects of update frequency and irrelevant data.}    
    \label{fig:parameter-analysis}
    \vspace{-0.5cm}
\end{figure}
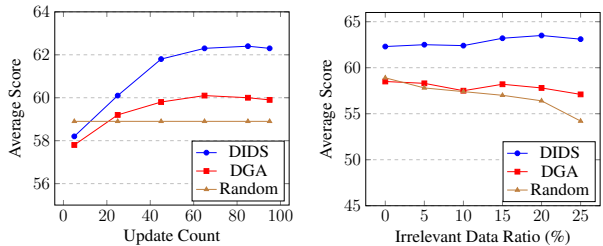

%% file: figure/fig-q5.tex
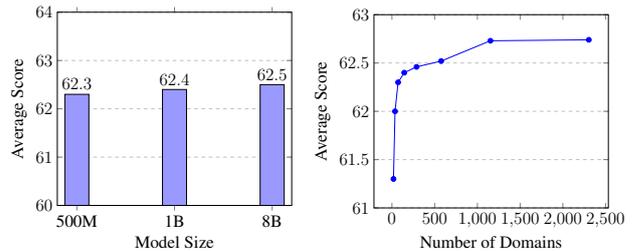
\begin{figure}[t]
    \subfigure[Impact of Proxy Model Size]{
    \begin{minipage}[t]{0.5\linewidth}
        \centering
        \begin{tikzpicture}[scale=0.45]
        \begin{axis}[
            xlabel={Model Size},
            ylabel={Average Score},
            symbolic x coords={500M,1B,8B},
            xtick=data,
            ymajorgrids=true,
            grid style=dashed,
            ymin=60,
            ymax=64,
            ybar,
            bar width=20pt,
            font=\Large,
            nodes near coords,
            nodes near coords align={vertical},
        ]
        \addplot[fill=blue!40] coordinates {
            (500M,62.3)
            (1B,62.4)
            (8B,62.5)
        };
        \end{axis}
        \end{tikzpicture}
    \end{minipage}
    }%
    \subfigure[Impact of Domain Count]{
    \begin{minipage}[t]{0.5\linewidth}
        \centering
        \begin{tikzpicture}[scale=0.45]
        \begin{axis}[
            xlabel={Number of Domains},
            ylabel={Average Score},
            legend style={at={(0.98,0.02)}, anchor=south east},
            ymajorgrids=true,
            grid style=dashed,
            ymin=61,
            ymax=63,
            font=\Large,
        ]
        \addplot[thick,mark=*,blue] coordinates {
            (18,61.3)
            (36,62.0)
            (72,62.3)
            (144,62.4)
            (288,62.46)
            (576,62.52)
            (1152,62.73)
            (2304,62.74)
        };
        \end{axis}
        \end{tikzpicture}
    \end{minipage}
    }
    \vspace{-0.3cm}
    \caption{Effects of model size and domain count.}
    \label{fig:model-analysis}
    \vspace{-0.5cm}
\end{figure}

%% file: figure/t1.tex
\begin{table}[t]
\footnotesize
\centering
\begin{tabular}{lrr}
\toprule
\textbf{Dataset} & \textbf{Samples} & \textbf{Percentage (\%)} \\
\midrule
Tulu 3 Persona MATH & 149,960 & 16.0 \\
Evol CodeAlpaca & 107,276 & 11.4 \\
FLAN v2 & 89,982 & 9.6 \\
NuminaMath-TIR & 64,312 & 6.8 \\
Tulu 3 Persona GSM & 49,980 & 5.3 \\
Tulu 3 WildGuardMix & 50,000 & 5.3 \\
Tulu 3 WildJailbreak & 50,000 & 5.3 \\
Tulu 3 Persona Python & 34,999 & 3.7 \\
Tulu 3 Persona IF & 29,980 & 3.2 \\
Tulu 3 Persona Algebra & 20,000 & 2.1 \\
CoCoNot & 10,983 & 1.2 \\
No Robots & 9,500 & 1.0 \\
OpenAssistant Guanaco & 7,132 & 0.8 \\
TableGPT & 5,000 & 0.5 \\
Tulu 3 Hardcoded & 240 & 0.03 \\
Aya & 100,000 & 10.6 \\
WildChat GPT-4 & 100,000 & 10.6 \\
SciRIFF & 10,000 & 1.1 \\
\midrule
\textbf{Total} & 939,344 & 100.0 \\
\bottomrule
\end{tabular}
\caption{Distribution of training data across different sources.}
\label{tab:data_dist}
\end{table}

%% file: figure/mixtral_tulu.tex
\begin{table*}[thbp]
\centering
\small
\begin{tabular}{l|ccc|cc|c|cc|c|c}
\toprule
\multirow{2}{*}{Method} & \multicolumn{3}{c|}{Reasoning} & \multicolumn{2}{c|}{Mathematics} & Instruction & \multicolumn{2}{c|}{Commonsense} & Truthfulness & \multirow{2}{*}{Average} \\
& BBH & BoolQ & GSM8K & MathQA & IFEval & MMLU & PIQA & PubMedQA & TruthfulQA & \\
\midrule
\multicolumn{11}{c}{Multi-task Optimization} \\
\midrule
Mixtral-7B & 56.0 & 84.7 & 36.9 & 13.2 & 36.6 & 61.9 & 81.6 & 77.8 & 41.3 & 54.4 \\
+ Full Data (929k) & 61.0 & 87.8 & 52.5 & 14.0 & 63.2 & 62.0 & 81.8 & 79.5 & 42.0 & 60.4 \\
+ Random (100k) & 60.2 & 86.7 & 47.8 & 9.5 & 55.3 & 61.8 & 82.2 & 78.5 & 43.0 & 58.3 \\
+ Uniform (100k) & 59.1 & 85.3 & 46.5 & 9.8 & 55.0 & 61.5 & 81.8 & 77.5 & 42.8 & 57.7 \\
+ Doremi (100k) & 60.5 & 86.5 & 48.0 & 15.0 & 56.5 & 62.0 & 82.0 & 79.0 & 46.5 & 59.5 \\
+ Velocitune (100k) & 60.2 & 86.2 & 46.0 & 14.8 & 56.0 & 61.8 & 81.9 & 78.8 & 45.8 & 59.1 \\
+ Doge (100k) & 60.8 & 86.8 & 47.0 & 15.3 & 57.8 & 62.1 & 82.3 & 79.6 & 47.2 & 59.9 \\
+ DGA (100k) & 60.0 & 86.3 & 48.0 & 15.8 & 53.5 & 62.2 & 82.0 & 77.0 & 44.5 & 58.8 \\
+ DIDS (100k) & \textbf{61.5} & \textbf{87.0} & \textbf{48.5} & \textbf{17.8} & \textbf{60.0} & \textbf{62.4} & \textbf{82.5} & \textbf{81.0} & \textbf{50.5} & \textbf{61.2} \\
\midrule
\multicolumn{11}{c}{Single-task Optimization} \\
\midrule
+ Doremi (100k) & 61.8 & 86.8 & 50.0 & 15.8 & 57.5 & 62.3 & 82.8 & 79.8 & 47.0 & 60.4 \\
+ Velocitune (100k) & 61.0 & 86.5 & 49.5 & 15.5 & 57.0 & 62.0 & 82.3 & 79.5 & 46.5 & 60.0 \\
+ Doge (100k) & 61.3 & 87.0 & 50.2 & 16.0 & 57.8 & 62.4 & 82.5 & 80.0 & 47.5 & 60.5 \\
+ DGA (100k) & 61.7 & 86.8 & 51.0 & 16.7 & 58.8 & 62.6 & 82.6 & 79.8 & 48.0 & 60.9 \\
+ DIDS (100k) & \textbf{62.5} & \textbf{87.5} & \textbf{52.0} & \textbf{18.5} & \textbf{62.0} & \textbf{63.0} & \textbf{83.0} & \textbf{82.0} & \textbf{52.0} & \textbf{62.5} \\
\bottomrule
\end{tabular}
\caption{Performance comparison of Mixtral-7B model trained on Tulu-v3 dataset under different sampling strategies.}
\label{tab:mixtral_tulu}
\end{table*}

%% file: figure/llama_openhermes.tex
\begin{table*}[thbp]
\centering
\small
\begin{tabular}{l|ccc|cc|c|cc|c|c}
\toprule
\multirow{2}{*}{Method} & \multicolumn{3}{c|}{Reasoning} & \multicolumn{2}{c|}{Mathematics} & Instruction & \multicolumn{2}{c|}{Commonsense} & Truthfulness & \multirow{2}{*}{Average} \\
& BBH & BoolQ & GSM8K & MathQA & IFEval & MMLU & PIQA & PubMedQA & TruthfulQA & \\
\midrule
\multicolumn{11}{c}{Multi-task Optimization} \\
\midrule
Llama-3.1-8B & 62.5 & 81.8 & 48.9 & 15.7 & 18.5 & 64.7 & 81.1 & 75.8 & 28.5 & 53.1 \\
+ Full OH-2.5 (1000k) & 67.5 & 86.8 & 67.0 & 17.5 & 60.0 & 64.5 & 81.5 & 77.5 & 39.5 & 62.4 \\
+ Random (100k) & 66.8 & 85.0 & 60.2 & 12.5 & 49.0 & 64.2 & 82.0 & 76.8 & 34.0 & 59.0 \\
+ Uniform (100k) & 65.8 & 83.5 & 59.0 & 12.8 & 48.8 & 64.0 & 81.6 & 76.2 & 33.5 & 58.4 \\
+ Doremi (100k) & 67.2 & 85.5 & 61.5 & 17.0 & 50.5 & 64.5 & 82.0 & 77.5 & 38.0 & 60.4 \\
+ Velocitune (100k) & 67.0 & 85.2 & 60.0 & 16.8 & 50.0 & 64.3 & 81.8 & 77.2 & 37.5 & 60.0 \\
+ Doge (100k) & 67.5 & 85.8 & 61.0 & 17.5 & 52.0 & 64.6 & 82.1 & 78.0 & 39.0 & 60.8 \\
+ DGA (100k) & 66.8 & 85.2 & 61.8 & 17.8 & 46.5 & 64.7 & 81.9 & 76.5 & 36.5 & 59.7 \\
+ DIDS (100k) & \textbf{68.0} & \textbf{86.5} & \textbf{62.5} & \textbf{19.5} & \textbf{56.0} & \textbf{64.8} & \textbf{82.3} & \textbf{79.5} & \textbf{45.0} & \textbf{62.7} \\
\midrule
\multicolumn{11}{c}{Single-task Optimization} \\
\midrule
+ Doremi (100k) & 68.5 & 86.0 & 63.0 & 18.0 & 52.5 & 64.8 & 82.5 & 78.2 & 39.0 & 61.4 \\
+ Velocitune (100k) & 67.8 & 85.8 & 62.5 & 17.8 & 52.0 & 64.6 & 82.2 & 78.0 & 38.6 & 61.0 \\
+ Doge (100k) & 68.0 & 86.5 & 63.0 & 18.2 & 53.0 & 64.8 & 82.3 & 78.8 & 39.5 & 61.6 \\
+ DGA (100k) & 68.4 & 86.2 & 64.0 & 19.0 & 54.5 & 65.0 & 82.5 & 78.5 & 40.5 & 62.1 \\
+ DIDS (100k) & \textbf{69.0} & \textbf{87.2} & \textbf{65.5} & \textbf{21.0} & \textbf{58.5} & \textbf{65.5} & \textbf{82.8} & \textbf{80.5} & \textbf{46.5} & \textbf{64.1} \\
\bottomrule
\end{tabular}
\caption{Performance comparison of Llama-3.1-8B model trained on OpenHermes-2.5 dataset under different sampling strategies.}
\label{tab:llama_openhermes}
\end{table*}

%% file: figure/mixtral_openhermes.tex
\begin{table*}[thbp]
\centering
\small
\begin{tabular}{l|ccc|cc|c|cc|c|c}
\toprule
\multirow{2}{*}{Method} & \multicolumn{3}{c|}{Reasoning} & \multicolumn{2}{c|}{Mathematics} & Instruction & \multicolumn{2}{c|}{Commonsense} & Truthfulness & \multirow{2}{*}{Average} \\
& BBH & BoolQ & GSM8K & MathQA & IFEval & MMLU & PIQA & PubMedQA & TruthfulQA & \\
\midrule
\multicolumn{11}{c}{Multi-task Optimization} \\
\midrule
Mixtral-7B & 56.0 & 84.7 & 36.9 & 13.2 & 36.6 & 61.9 & 81.6 & 77.8 & 41.3 & 54.4 \\
+ Full OH-2.5 (1000k) & 59.8 & 87.9 & 64.2 & 14.4 & 45.8 & 62.0 & 82.6 & 76.6 & 50.5 & 60.4 \\
+ Random (100k) & 58.5 & 86.5 & 54.0 & 10.5 & 42.0 & 61.8 & 82.0 & 76.0 & 46.0 & 57.5 \\
+ Uniform (100k) & 57.8 & 86.0 & 52.5 & 10.8 & 41.8 & 61.7 & 81.8 & 75.5 & 45.5 & 57.0 \\
+ Doremi (100k) & 59.0 & 87.0 & 55.5 & 13.8 & 43.5 & 62.0 & 82.2 & 76.2 & 48.0 & 58.6 \\
+ Velocitune (100k) & 58.7 & 86.8 & 54.5 & 13.5 & 43.0 & 61.9 & 82.0 & 76.0 & 47.5 & 58.2 \\
+ Doge (100k) & 59.2 & 87.2 & 55.8 & 14.0 & 44.0 & 62.1 & 82.3 & 76.8 & 48.5 & 58.9 \\
+ DGA (100k) & 58.5 & 86.8 & 56.5 & 14.2 & 41.0 & 62.2 & 82.1 & 75.8 & 47.0 & 58.2 \\
+ DIDS (100k) & \textbf{60.0} & \textbf{87.5} & \textbf{58.0} & \textbf{15.8} & \textbf{45.0} & \textbf{62.3} & \textbf{82.5} & \textbf{77.5} & \textbf{52.0} & \textbf{60.1} \\
\midrule
\multicolumn{11}{c}{Single-task Optimization} \\
\midrule
+ Doremi (100k) & 60.5 & 87.5 & 58.0 & 14.5 & 44.5 & 62.2 & 82.8 & 77.0 & 49.0 & 59.6 \\
+ Velocitune (100k) & 60.0 & 87.2 & 57.5 & 14.2 & 44.0 & 62.0 & 82.5 & 76.8 & 48.5 & 59.2 \\
+ Doge (100k) & 60.2 & 87.8 & 58.2 & 14.8 & 44.8 & 62.3 & 82.7 & 77.2 & 49.5 & 59.7 \\
+ DGA (100k) & 60.8 & 87.5 & 59.5 & 15.0 & 46.0 & 62.5 & 82.9 & 77.0 & 50.0 & 60.1 \\
+ DIDS (100k) & \textbf{61.5} & \textbf{88.0} & \textbf{61.0} & \textbf{16.5} & \textbf{47.5} & \textbf{62.8} & \textbf{83.0} & \textbf{78.0} & \textbf{53.0} & \textbf{61.3} \\
\bottomrule
\end{tabular}
\caption{Performance comparison of Mixtral-7B model trained on OpenHermes-2.5 dataset under different sampling strategies.}
\label{tab:mixtral_openhermes}
\end{table*}

%% file: figure/full_ablation.tex
\begin{table*}[thbp]
\centering
\small
\begin{tabular}{l|ccc|cc|c|cc|c|c}
\toprule
\multirow{2}{*}{Variant} & \multicolumn{3}{c|}{Reasoning} & \multicolumn{2}{c|}{Mathematics} & Instruction & \multicolumn{2}{c|}{Commonsense} & Truthfulness & \multirow{2}{*}{Average} \\
& BBH & BoolQ & GSM8K & MathQA & IFEval & MMLU & PIQA & PubMedQA & TruthfulQA & \\
\midrule
DIDS (100k) & \textbf{68.3} & \textbf{86.9} & \textbf{59.0} & \textbf{20.5} & \textbf{55.6} & \textbf{64.9} & \textbf{82.2} & \textbf{80.4} & \textbf{43.0} & \textbf{62.3} \\
DIDS-GC & 67.7 & 85.7 & 58.2 & 19.7 & 53.0 & 64.4 & 81.8 & 78.9 & 40.1 & 61.1 \\
DIDS-FIM & 67.2 & 85.0 & 57.4 & 18.6 & 51.9 & 64.1 & 81.5 & 77.2 & 38.5 & 60.2 \\
DIDS-LT & 67.5 & 85.3 & 57.8 & 19.5 & 51.4 & 64.2 & 81.6 & 77.5 & 38.1 & 60.3 \\
\bottomrule
\end{tabular}
\caption{Comprehensive ablation study of DIDS across all downstream tasks. DIDS-GC replaces gradient-based clustering with BERT semantic clustering, DIDS-FIM removes the FIM-guided impact measurement, and DIDS-LT eliminates the loss trajectory and saturation consideration.}
\label{tab:full_ablation}
\end{table*}

%% file: figure/domain_weights.tex
\begin{table*}[t]
\centering
\small
\begin{tabular}{l|c|ccccccccccc}
\toprule
Domain ID & Method & 0\% & 10\% & 20\% & 30\% & 40\% & 50\% & 60\% & 70\% & 80\% & 90\% & 100\% \\
\midrule
D023 & DIDS & 0.4 & 0.5 & 0.7 & 0.9 & 1.2 & 1.6 & 1.9 & 2.3 & 2.5 & 2.7 & 2.8 \\
 & DGA & 0.4 & 0.4 & 0.5 & 0.6 & 0.7 & 0.9 & 1.0 & 1.1 & 1.2 & 1.3 & 1.3 \\
\midrule
D045 & DIDS & 0.4 & 0.3 & 0.2 & 0.2 & 0.1 & 0.1 & 0.1 & 0.0 & 0.0 & 0.0 & 0.0 \\
 & DGA & 0.4 & 0.3 & 0.3 & 0.2 & 0.2 & 0.2 & 0.2 & 0.1 & 0.1 & 0.1 & 0.1 \\
\midrule
D078 & DIDS & 0.4 & 0.6 & 0.8 & 1.0 & 1.5 & 1.2 & 1.2 & 1.3 & 1.3 & 1.4 & 1.3 \\
 & DGA & 0.4 & 0.5 & 0.6 & 0.7 & 0.8 & 0.8 & 0.9 & 0.4 & 0.6 & 0.3 & 0.4 \\
\midrule
D102 & DIDS & 0.4 & 0.7 & 0.9 & 1.2 & 1.4 & 1.5 & 1.6 & 1.7 & 1.7 & 1.7 & 1.7 \\
 & DGA & 0.4 & 0.5 & 0.6 & 0.7 & 0.8 & 0.8 & 0.9 & 1.0 & 1.0 & 1.0 & 1.0 \\
\midrule
D129 & DIDS & 0.4 & 0.2 & 0.1 & 0.0 & 0.0 & 0.0 & 0.0 & 0.0 & 0.0 & 0.0 & 0.0 \\
 & DGA & 0.4 & 0.3 & 0.2 & 0.2 & 0.1 & 0.1 & 0.1 & 0.1 & 0.1 & 0.0 & 0.0 \\
\midrule
D147 & DIDS & 0.4 & 0.3 & 0.2 & 0.2 & 0.1 & 0.1 & 0.1 & 0.1 & 0.1 & 0.0 & 0.0 \\
 & DGA & 0.4 & 0.4 & 0.3 & 0.3 & 0.3 & 0.2 & 0.2 & 0.2 & 0.2 & 0.2 & 0.1 \\
\midrule
D175 & DIDS & 0.4 & 0.5 & 0.5 & 0.6 & 0.6 & 0.7 & 0.7 & 0.7 & 0.7 & 0.8 & 0.8 \\
 & DGA & 0.4 & 0.5 & 0.5 & 0.5 & 0.6 & 0.6 & 0.6 & 0.6 & 0.6 & 0.6 & 0.6 \\
\midrule
D198 & DIDS & 0.4 & 0.7 & 1.0 & 1.3 & 1.5 & 1.7 & 1.8 & 1.9 & 1.9 & 1.9 & 1.9 \\
 & DGA & 0.4 & 0.6 & 0.7 & 0.8 & 0.9 & 0.9 & 1.0 & 1.0 & 1.0 & 1.0 & 1.0 \\
\midrule
D221 & DIDS & 0.4 & 0.2 & 0.1 & 0.0 & 0.0 & 0.0 & 0.0 & 0.0 & 0.0 & 0.0 & 0.0 \\
 & DGA & 0.4 & 0.3 & 0.2 & 0.1 & 0.1 & 0.1 & 0.0 & 0.0 & 0.0 & 0.0 & 0.0 \\
\midrule
D244 & DIDS & 0.4 & 0.4 & 0.5 & 0.4 & 0.4 & 0.4 & 0.4 & 0.5 & 0.4 & 0.4 & 0.4 \\
 & DGA & 0.4 & 0.4 & 0.4 & 0.4 & 0.4 & 0.4 & 0.4 & 0.4 & 0.4 & 0.4 & 0.4 \\
\bottomrule
\end{tabular}
\caption{Comparison of domain weight evolution between DIDS and DGA across training progress (from 0\% to 100\% completion).}
\label{tab:domain_weights}
\end{table*}

%% file: figure/grid_search.tex
\begin{table}[t]
\centering
\scriptsize
\begin{tabular}{lccc}
\toprule
\textbf{Mixing Strategy} & \textbf{GSM8K} & \textbf{HumanEval} & \textbf{MT-Bench} \\
\midrule
Mix[(code,math), 1 general] & 47.53 & 14.63 & 5.76 \\
Mix[(code,math), 1/4 general] & 48.44 & 15.85 & 5.73 \\
Mix[(code,math), 1/16 general] & 47.99 & 15.24 & 5.27 \\
Mix[(code,math), 1/64 general] & 47.23 & 14.63 & 5.16 \\
Mix[(code,math), 1/256 general] & 48.52 & 16.46 & 4.69 \\
\midrule
Mix[1(code,math), general] & 47.53 & 14.63 & 5.76 \\
Mix[1/4(code,math), general] & 41.31 & 10.97 & 5.81 \\
Mix[1/16(code,math), general] & 33.20 & 11.58 & 5.76 \\
Mix[1/64(code,math), general] & 25.17 & 12.19 & 5.84 \\
Mix[1/256(code,math), general] & 16.52 & 9.14 & 5.82 \\
\midrule
DIDS (dynamic) & 52.21 & 18.05 & 5.88 \\
\bottomrule
\end{tabular}
\caption{Performance with different static mixing ratios between specialized and general domains.}
\label{tab:grid_search}
\end{table}

%% file: figure/alg.tex
\begin{algorithm*}[t]
\caption{Domain Impact-aware Data Sampling (DIDS)}
\label{alg:dids}
\KwIn{Training dataset $\mathcal{D}$; Downstream tasks $\mathcal{S}$; Proxy model $f'$; Number of domains $k$; Update interval $\tau$; EMA coefficient $\beta$; Training steps $T$}
\KwOut{Domain sampling probabilities $\mathbf{p}_t$}

// Initialize sampling probabilities uniformly\;
$\mathbf{p}_0 \leftarrow [1/k,...,1/k]$\;

// Domain repartition based on gradients\;
$G \leftarrow \emptyset$\;
\ForEach{$x_i \in \mathcal{D}$}{
    $g_i \leftarrow \nabla \ell(f', x_i)$ \;
    $g_i \leftarrow \text{TopK}(g_i)$ \;
    $\tilde{g}_i \leftarrow R^T g_i$\;
    $G \leftarrow G \cup \{\tilde{g}_i\}$\;
}
$\{D_1,...,D_k\} \leftarrow \text{KMeans}(G, k)$\;

\For{$t \leftarrow 1$ \KwTo $T$}{
    \If{$t \bmod \tau = 0$}{
        // Compute domain impact matrix\;
        \ForEach{$D_i \in \{D_1,...,D_k\}$}{
            \ForEach{$S_j \in \{S_1,...,S_m\}$}{
                $\Delta \leftarrow \nabla \ell_{S_j} - \nabla \ell_{D_i}$\;
                $F \leftarrow \mathbb{E}[\nabla\log p(\theta) \odot \nabla\log p(\theta)]$\;
                $I(D_i,S_j) \leftarrow \frac{1}{2}\Delta^T F \Delta$\;
            }
        }
        
        // Compute future potential\;
        \ForEach{$S_j \in \{S_1,...,S_m\}$}{
            Fit $L(t) = ae^{-bt} + c$ using loss history $\{L_1(S_j),...,L_t(S_j)\}$\;
            $L_p(S_j) \leftarrow L_t(S_j) - L(t + \tau)$\;
            $\Delta L(S_j) \leftarrow L_{t-1}(S_j) - L_t(S_j)$\;
        }
        
        // Update sampling probabilities\;
        \ForEach{$D_i \in \{D_1,...,D_k\}$}{
            $U(D_i) \leftarrow \sum_j \frac{I(D_i,S_j) \cdot (\Delta L(S_j) + L_p(S_j))}{p_{t-1,i}}$\;
            $\hat{p}_{t,i} \leftarrow \text{softmax}(U(D_i))$\;
            $p_{t,i} \leftarrow \beta p_{t-1,i} + (1-\beta)\hat{p}_{t,i}$\;
        }
        $\mathbf{p}_t \leftarrow \mathbf{p}_t / \sum_i p_{t,i}$\;
    }
    Sample batch $\mathcal{B}_t$ according to $\mathbf{p}_t$\;
    Update model parameters $\theta$ using $\mathcal{B}_t$\;
}
\Return{$\mathbf{p}_t$}
\end{algorithm*}